\documentclass[10pt,twocolumn,letterpaper]{article}

\usepackage[pagenumbers]{cvpr} 

\usepackage{graphicx}
\usepackage{amsmath}
\usepackage{amssymb}
\usepackage{booktabs}
\usepackage{float}
\usepackage{multirow}
\usepackage[accsupp]{axessibility}

\usepackage[pagebackref,breaklinks,colorlinks]{hyperref}
\usepackage[capitalize]{cleveref}
\crefname{section}{Sec.}{Secs.}
\Crefname{section}{Section}{Sections}
\Crefname{table}{Table}{Tables}
\crefname{table}{Tab.}{Tabs.}

\def\cvprPaperID{1714}
\def\confName{CVPR}
\def\confYear{2022}

\begin{document}

\title{Learning Deep Implicit Functions for 3D Shapes with Dynamic Code Clouds}

\author{
Tianyang Li\textsuperscript{1}, Xin Wen\textsuperscript{1,2}, Yu-Shen Liu\textsuperscript{1}\thanks{The corresponding author is Yu-Shen Liu. This work was supported by National Key R\&D Program of China (2018YFB0505400, 2020YFF0304100), the National Natural Science Foundation of China (62072268), and in part by Tsinghua-Kuaishou Institute of Future Media Data.}, Hua Su\textsuperscript{3}, Zhizhong Han\textsuperscript{4}\\
\textsuperscript{1} School of Software, BNRist, Tsinghua University, Beijing, China\\
\textsuperscript{2} JD Logistics, Beijing, China\\
\textsuperscript{3} Kuaishou Technology, Beijing, China\\
\textsuperscript{4} Department of Computer Science, Wayne State University, Detroit, USA\\
{\small lity20@mails.tsinghua.edu.cn \hspace{1mm} wenxin16@jd.com \hspace{1mm} liuyushen@tsinghua.edu.cn}\\
{\small shlw@kuaishou.com \hspace{1mm} h312h@wayne.edu}
}


\maketitle

\begin{abstract}
Deep Implicit Function (DIF) has gained popularity as an efficient 3D shape representation. To capture geometry details, current methods usually learn DIF using local latent codes, which discretize the space into a regular 3D grid (or octree) and store local codes in grid points (or octree nodes). Given a query point, the local feature is computed by interpolating its neighboring local codes with their positions. However, the local codes are constrained at discrete and regular positions like grid points, which makes the code positions difficult to be optimized and limits their representation ability. To solve this problem, we propose to learn DIF with Dynamic Code Cloud, named DCC-DIF. Our method explicitly associates local codes with learnable position vectors, and the position vectors are continuous and can be dynamically optimized, which improves the representation ability. In addition, we propose a novel code position loss to optimize the code positions, which heuristically guides more local codes to be distributed around complex geometric details. In contrast to previous methods, our DCC-DIF represents 3D shapes more efficiently with a small amount of local codes, and improves the reconstruction quality. Experiments demonstrate that DCC-DIF achieves better performance over previous methods. Code and data are available at \href{https://github.com/lity20/DCCDIF}{https://github.com/lity20/DCCDIF}.
\end{abstract}

\section{Introduction}
\label{sec:introduction}

Learning 3D shape representation is important for many downstream applications in 3D computer vision \cite{2DProjectionMatching,Han2020Reconstructing3S,Han2020DRWRAD,Han2021HierarchicalVP,Han20193D2SeqViewsAS,han2019SeqViews2SeqLabels,Han2019MultiAnglePC}. Explicit 3D representations such as meshes, voxels and point clouds have been widely used in various tasks \cite{Kato2018Neural3M,Qi2017PointNetDL,Qi2017PointNetDH,Maturana2015VoxNetA3,xiang2021snowflakenet,wen2021pmp,Wen2020PointCC,wen2021c4c,Wen2020Point2SpatialCapsuleAF,Wen2020CFSISSS}. Recently, deep implicit function (DIF) \cite{Mescheder2019OccupancyNL,Park2019DeepSDFLC,Chen2019LearningIF,Xu2019DISNDI,NeuralPull,Ma2022RS,Ma2022SR} has received more popularity as an efficient 3D shape representation, which learns latent codes of 3D shapes by predicting a signed distance or inside/outside for each query point. Different from explicit 3D representations, DIF can be stored compactly and learn shape priors by the network. Besides, it is simple and natural to use DIF in learning-based tasks because of its differentiable ability.

\begin{figure}[t]
    \centering
    \includegraphics[scale=0.45]{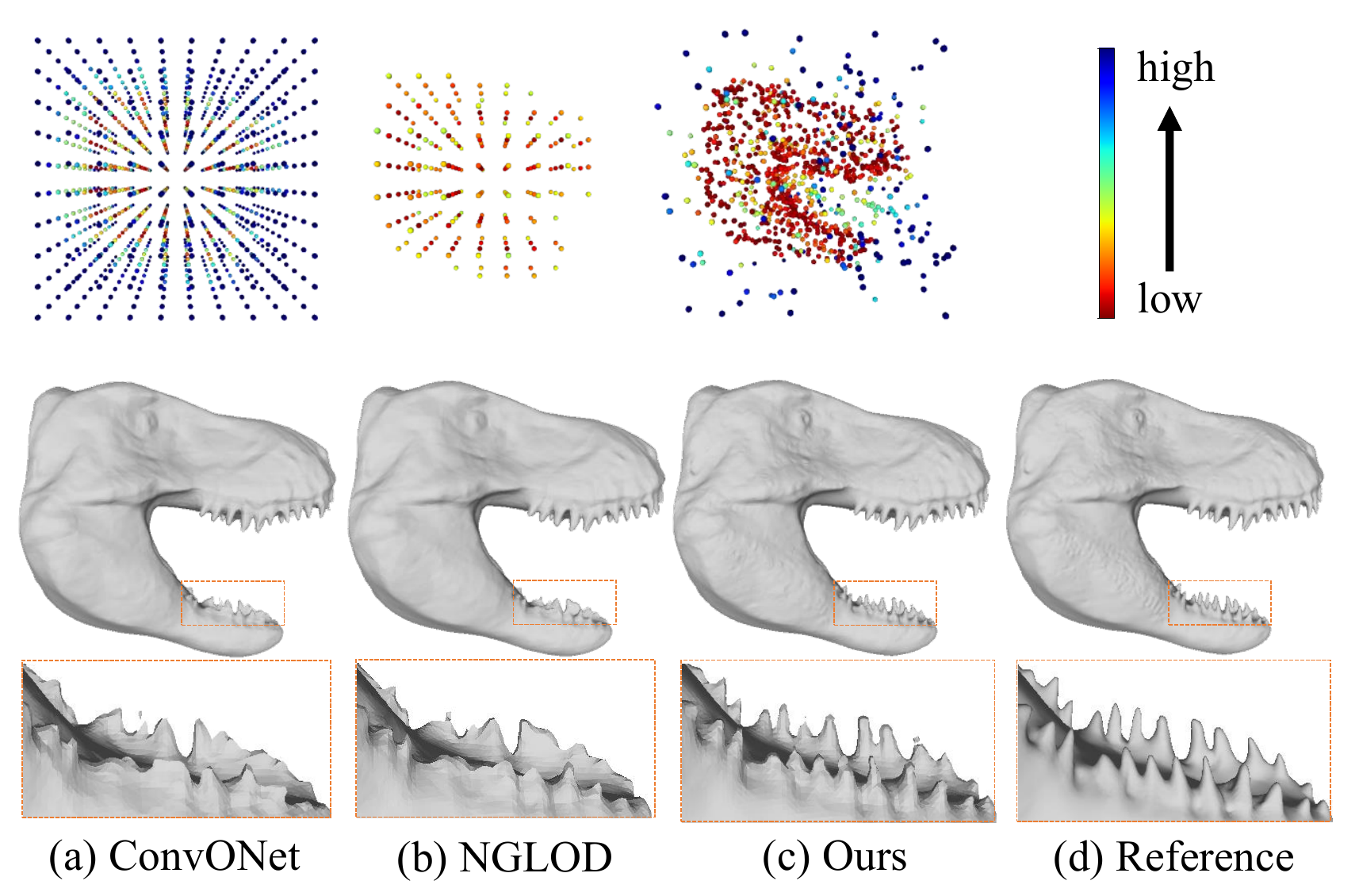}
    \caption{Illustration comparison between our method and other methods. In (a), we select the grid-based DIF (ConvONet \cite{Peng2020ECCV}). In (b), we show the octree-based DIF (NGLOD \cite{Takikawa2021NeuralGL}). (c) is our DCC-DIF. (d) is Reference.
    The first row shows the code positions of different methods, where the warmer color indicates the local codes are closer to the surface. Compared with other methods, in which code positions are discrete and regular, our code positions are continuous and more flexible. The second row shows the reconstruction results, where our method can reconstruct highly detailed geometry of complex shapes, like teeth.
}
    \label{fig:intro}
\end{figure}

Previous DIF approaches \cite{Mescheder2019OccupancyNL,Park2019DeepSDFLC,Chen2019LearningIF} encode the entire 3D shape into a single global latent code through the auto-encoder or auto-decoder \cite{Tan1995ReducingDD} framework, which leads to information loss of local regions. As a result, those approaches can not capture local geometry details well and struggle to represent complex shapes. To address this problem, some methods \cite{Chabra2020DeepLS,Jiang2020LocalIG} divide the 3D space into small local volumes and assign each volume with a latent code. Then each local volume is reconstructed separately and all volumes are combined together to get the final reconstruction. Since small local volumes contain simple shapes and common patterns are shared among volumes, these methods can represent 3D shapes with high accuracy and generalize to different shapes. Similarly, some approaches \cite{Genova2020LocalDI, Genova2019LearningST} decompose 3D shapes into local parts, each of which is associated with a latent code for learning local details. 
On the other hand, more recent methods discretize the space into a regular 3D grid \cite{Peng2020ECCV,Chibane2020ImplicitFI,Chen_2021_ICCV} (or octree) \cite{Martel2021ACORNAC,Takikawa2021NeuralGL} and store the local codes in the grid points (or octree nodes). Given a query point in 3D space, the local feature is computed by interpolating its neighboring local codes with their positional weights. Next, the local feature is fed into a decoder to predict a signed distance or inside/outside. As the resolution of grids or depth of octree increases, these methods achieve the state-of-the-art results in several shape reconstruction
tasks. However, the increase of resolution or depth will result in a significant growth of the number of local codes. Moreover, the local codes in these methods are constrained at discrete and regular positions like grid points or octree nodes, which makes the code positions difficult to be optimized \cite{Peng2020ECCV,Chen_2021_ICCV,Chibane2020ImplicitFI,Takikawa2021NeuralGL,Martel2021ACORNAC} and limits the representation ability.

\begin{figure}[t]
    \centering
    \includegraphics[scale=0.4]{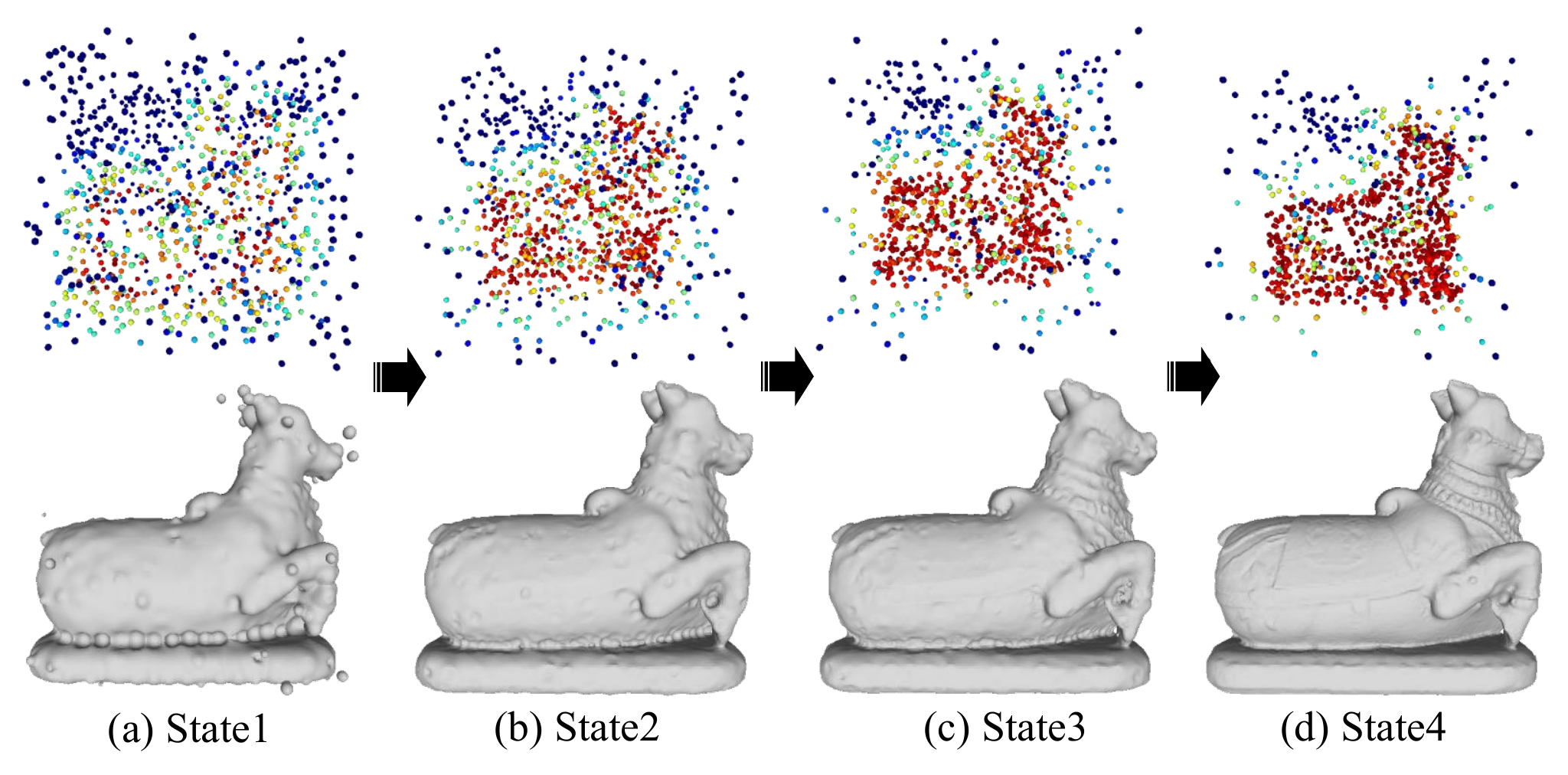}
    \caption{\textbf{Illustration of moving code positions during optimization.} Our code positions are dynamically updated during optimization, which makes 3D shape representation more efficiently. We typically show four states during optimization, where the first and second rows display code positions and reconstructions, respectively. Initial and final states are shown in (a) and (d), respectively, while (b) and (c) show two intermediate states.}
    \label{fig:optimization_process}
\end{figure}

To address the above-mentioned problems, we propose a novel method to learn DIF for 3D shape with Dynamic Code Cloud, named DCC-DIF. Specifically, we represent a 3D shape with a set of local latent codes, each of which is explicitly associated with a learnable position vector. Using these position vectors, the local feature for a query point is computed by interpolating local codes with their positional weights, which are computed using the distances relative to this query point. 
In contrast with previous local DIF methods \cite{Chibane2020ImplicitFI,Chen_2021_ICCV,Peng2020ECCV,Takikawa2021NeuralGL}, which store the local codes in discrete and regular grids, the positions of local codes used in our method are continuous and flexible.
Specially, our code positions can be dynamically optimized, where the position vectors is learnable and can be updated by back-propagation and gradient descent. Therefore, we name our method \emph{Dynamic Code Cloud} (DCC), as shown in \cref{fig:intro} and \cref{fig:optimization_process}.
In addition, we design a novel \emph{Code Position} (CP) loss to optimize the positions of local codes, where more local codes are heuristically guided to distribute around complex geometric details. With the help of CP loss, our method can represent 3D shapes more efficiently with a small amount of local codes. As a result, when using the same number of local codes as previous methods, our method achieves better results and reconstructs highly detailed geometry of 3D shapes. 
Our main contributions are summarized as follows.
\begin{itemize}
    \item We propose a novel DCC-DIF to learn deep implicit function of 3D shapes. Compared with previous methods which limit the local codes at discrete and regular grid points, the code positions in DCC-DIF are continuous and can be dynamically optimized, which improves the representation ability.
    \item We further propose a novel code position (CP) loss to optimize the positions of local codes, so that more local codes are distributed around complex geometric details. With the help of CP loss, our DCC-DIF can represent 3D shapes with higher quality and efficiency.
    \item Compared to previous methods, our method can achieve better accuracy with fewer number of local codes when reconstructing highly detailed geometry of 3D shapes. Experiments demonstrate our DCC-DIF can achieve the state-of-the-art results.
\end{itemize}

\section{Related Work}
\label{sec:related_work}


The recent emerged implicit representation research has drawn a growing attention in 3D computer vision. Compared with the previous explicit representation based methods (e.g. voxel \cite{Maturana2015VoxNetA3}, mesh \cite{Kato2018Neural3M} and point cloud \cite{Qi2017PointNetDL,Qi2017PointNetDH}), the implicit methods can represent 3D shapes at arbitrary resolution. In this paper, we take the advantage of implicit 3D representation and focus on the task of reconstructing high-quality 3D signed distance functions (SDF). Relevant work of this area can be roughly divided into two categories, which are global DIF methods and the local DIF methods.

\paragraph{Global DIF methods.} 
For the previous global DIF methods, a common practice is to take the benefit from the traditional implicit representation methods, and integrate them into the deep learning based framework.
Typical method like DeepSDF \cite{Park2019DeepSDFLC} implicitly represents a 3D shape by its zero-level set. It optimizes a global latent code for each 3D shape, and predicts signed distances to the shape surface for sampled points using a decoder. On the other hand, OccNet \cite{Mescheder2019OccupancyNL} represents the surface of a shape by decision boundary of a deep neural network. It leverages an auto-encoder framework to predict inside/outside values for sampled points in 3D space. Following pioneers, some recent emerged methods have further improved the frontier of DIF research. For example, Duan et al. \cite{duan2020curriculum} learn DIF by a curriculum strategy, and Zheng et al. \cite{Zheng2021DeepIT} develop the deformation based method to predict DIF from shape templates. However, the problem is that these methods are still hard to  preserve the details of local surfaces, due to the fixed dimensionality of single global code.

\paragraph{Local DIF methods.} To overcome the limitation of global DIF methods, the local DIF methods have been developed to learn 3D shapes at more detailed geometric level. 
For example, LIG and DeepLS \cite{Jiang2020LocalIG,Chabra2020DeepLS} divide shapes/scenes into volumes, where each volume is independently reconstructed with an assigned latent code. After that, all volumes are combined together to get the final reconstruction. SIF, LDIF and PatchNets \cite{Genova2019LearningST,Genova2020LocalDI,Tretschk2020PatchNetsPG} decompose shapes into local patches and represent each patch using a latent code. More recently, IMLSNets \cite{Liu2021DeepIM} adapts implicit moving least squares surface formulation for learning based method. ConvONet \cite{Peng2020ECCV} builds 3D grids or 2D grids on each axes plane and stores a latent code in each grid point. Then given a query point in 3D space, the positions of this point and its neighboring grid points are leveraged to interpolate the stored latent codes into a vector. IF-Nets \cite{Chibane2020ImplicitFI} constructs hierarchical latent grids with different resolutions to capture local geometric information of different scales. Similarly, MDIF \cite{Chen_2021_ICCV} also constructs hierarchical latent grids. Moreover, it sets top level latent grids to be a global latent code and connects latent codes between different levels by transposed convolutions \cite{Kim2018MovingTC} and concatenations, which enables it to do global operations like completion. However, a high grid resolution is required for latent grids based methods to achieve good results, which leads to cubic growth of number of latent codes. NGLOD \cite{Takikawa2021NeuralGL} leverages the sparse octree instead of uniform grids to reduce the number of latent codes, and achieves state-of-the-art reconstruction accuracy. ACORN \cite{Martel2021ACORNAC} also adopts the octree, and the structure of octree can be adjusted during optimization. However, this step is non-differentiable and needs to solve an integer linear program problem. These grids or octree-based methods constrain that latent codes are located at discrete and regular positions like grid points or octree nodes. And code positions are static \cite{Takikawa2021NeuralGL,Chen_2021_ICCV,Chibane2020ImplicitFI,Peng2020ECCV} or non-trivial to be optimized \cite{Martel2021ACORNAC}. In contrast, positions of latent codes in our method are flexible and continuous. And we dynamically optimize code positions by back-propagation and gradient descent, which is more efficient than solving an integer linear program problem.

\begin{figure}[t]
    \centering
    \includegraphics[scale=0.62]{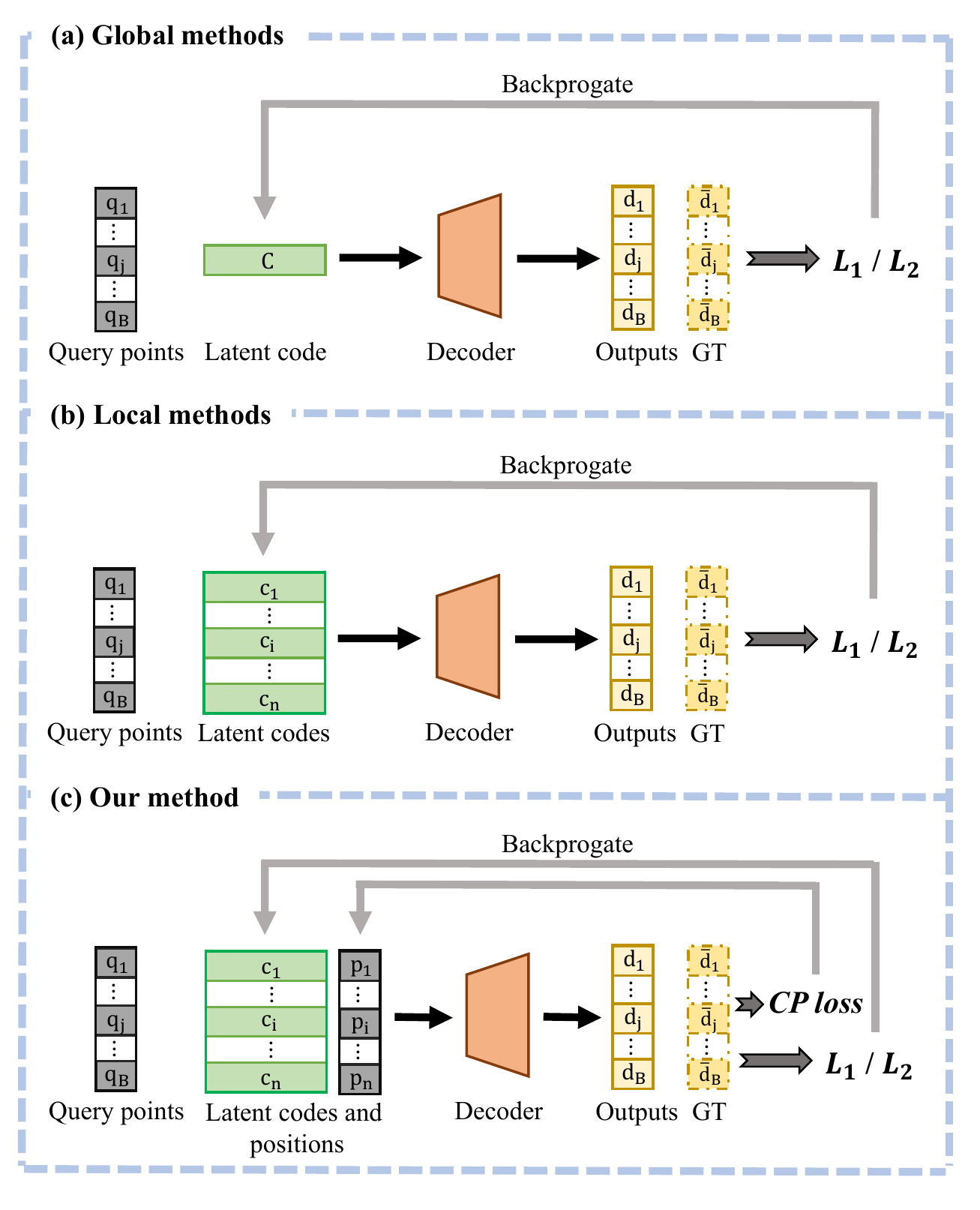}
    \caption{Illustration comparison between architecture of our method and other methods. We show the overall architecture of global DIF methods in (a), local DIF methods in (b), and our method in (c).}
    \label{fig:network}
\end{figure}

\section{Method}
\label{sec:method}

Our goal is to design a flexible 3D shape representation which can efficiently fit a single shape or reconstruct 3D datasets with high quality. \cref{fig:network} illustrates the overall architecture of our method and differences among global, local and our method. To represent a 3D shape, as shown in \cref{fig:network}(a), global methods leverage a single global latent code, and optimize the latent code by minimizing errors between outputs and the ground truth. Local methods replace the global latent code with a set of local latent codes, as shown in \cref{fig:network}(b). In our method, as shown in \cref{fig:network}(c), we explicitly assign a position vector to each local code, which indicates the $(x,y,z)$ coordinates of corresponding local code in 3D space, and a novel CP loss is further proposed to optimize position vectors.
In this section, we firstly introduce background knowledge about neural signed distance functions (SDF) in \cref{sec:deep_implicit_function}. Then the design of our method is explained in detail in \cref{sec:dynamic_code_cloud}. Next, we present our novel CP loss in \cref{sec:code_position_loss}. And lastly we describe the training process in \cref{sec:training}.

\subsection{Deep Implicit Function}
\label{sec:deep_implicit_function}

There are different approaches for deep implicit functions to represent surfaces. Mainstream approaches include occupancy functions \cite{Mescheder2019OccupancyNL} and signed distance functions (SDF) \cite{Park2019DeepSDFLC}. 
In this paper, we follow the paradigm of SDF. SDF can be formulated as $f:\mathbb{R}^3 \rightarrow \mathbb{R}$, and $d = f(\mathbf{x})$ is the shortest signed distance from a query point $\mathbf{x}$ to the surface of underlying 3D shape. And the sign determines whether it is inside or outside the 3D shape. Thus, the surface of a 3D shape is the zero level-set of SDF,  denoted as
\begin{equation}
    \mathcal{S} = \{ \mathbf{x} \in \mathbb{R}^3 \mid f(\mathbf{x}) = 0 \}.
    \label{eq:SDF}
\end{equation}

Learning-based SDF usually encodes a 3D shape into a single global latant code or local latent codes. And a multi-layer perceptron is leveraged as the decoder, which takes latent codes and query points as input and predicts signed distances. Using sampled query points as training data and the corresponding ground truth signed distances as supervision, the latent codes and network parameters are optimized by minimizing the errors between predicted and ground truth signed distances. After SDF is learned, the Marching Cubes algorithm \cite{Lorensen1987MarchingCA} is usually applied to extract an isosurface and outputs a mesh for rendering or visualization.

\subsection{Dynamic Code Cloud}
\label{sec:dynamic_code_cloud}

In our method, we leverage an auto-decoder framework \cite{Tan1995ReducingDD}. And we learn DIF using a novel Dynamic Code Cloud (DCC-DIF). In \cref{fig:network}, we show the overall architecture of our method and the differences between our method and the compared methods. We represent a 3D shape using a set of latent codes and the corresponding code positions, which are denoted by a matrix $\mathbf{C} \in \mathbb{R}^{n \times m}$ and a matrix $\mathbf{P} \in \mathbb{R}^{n \times 3}$, respectively. Here, $n$ indicates the number of local codes we used, and $m$ indicates the dimension of local codes. Each row of $\mathbf{C}$ is a latent code $\mathbf{c}_i \in \mathbb{R}^m$, and each row of $\mathbf{P}$ is a position vector $\mathbf{p}_i \in \mathbb{R}^3$, where $1 \leq i \leq n$. Each $\mathbf{c}_i$ and $\mathbf{p}_i$ form a pair and $\mathbf{p}_i$ indicates the $(x,y,z)$ coordinate of corresponding latent code in 3D space.

Given a batch of query points $\mathbf{Q} \in \mathbb{R}^{B \times 3}$ in 3D space with the batch size $B$, in which each row is a query point $\mathbf{q}_{j} \in \mathbb{R}^3 \ (1 \leq j \leq B)$, we firstly obtain a distance matrix $\mathcal{D} \in \mathbb{R}^{B \times n}$ between query points $\mathbf{Q}$ and code positions $\mathbf{P}$, where each element of $\mathcal{D}_{ji}$ is computed as
\begin{equation}
    \mathcal{D}_{ji} = || \mathbf{q}_j - \mathbf{p}_i ||_2.
    \label{eq:distance}
\end{equation}
Then a weight matrix $\mathcal{W}$ is obtained based on $\mathcal{D}$, each element of which is computed as
\begin{equation}
    \mathcal{W}_{ji} = \frac{\mathcal{W}_{ji}^\prime}{\sum_{k=1}^{n}\mathcal{W}^\prime_{jk}},
    \label{eq:weight}
\end{equation}
where
\begin{equation}
    \mathcal{W}_{ji}^\prime = \frac{1}{\mathcal{D}_{ji}^3}.
    \label{eq:weight_}
\end{equation}

\begin{figure}[t]
    \centering
    \includegraphics[scale=0.4]{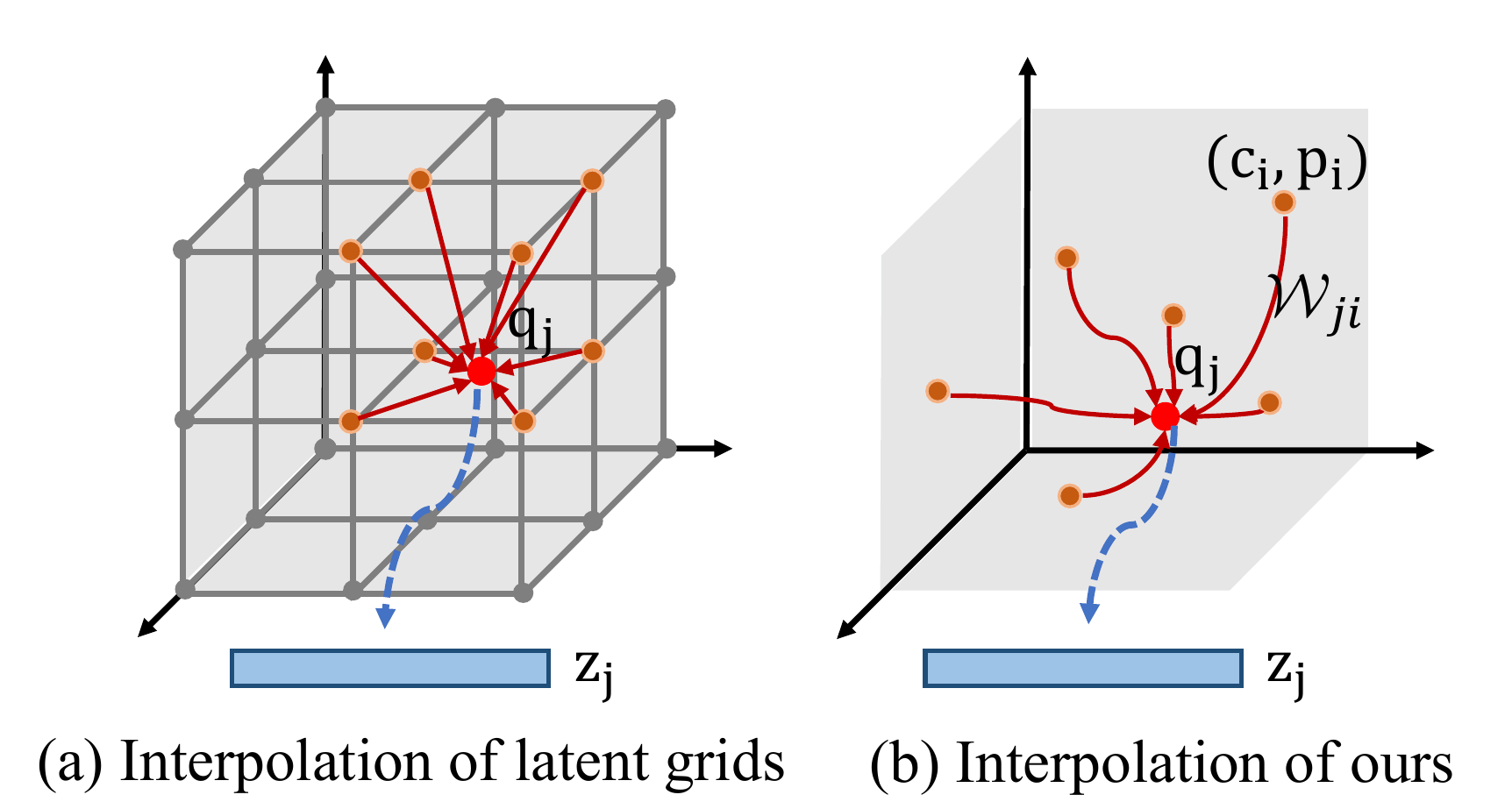}
    \caption{Illustration of interpolation. (a) Given a query point $\mathbf{q}_j$, previous grid-based methods \cite{Chibane2020ImplicitFI,Chen_2021_ICCV,Peng2020ECCV,Takikawa2021NeuralGL} leverage the position of $\mathbf{q}_j$ and its neighboring grid points to interpolate stored latent codes into a vector $\mathbf{z}_j$, where a trilinear interpolation is usually applied in these methods. (b) In our method, each latent code $\mathbf{c}_i$ is explicitly associated with a position vector $\mathbf{p}_i$ to indicate its position in 3D space, where the positions of both $\mathbf{q}_j$ and $\mathbf{p}_i$ are used to compute the weight $\mathcal{W}_{ji}$ which is used for interpolation.}
    \label{fig:interpolation}
\end{figure}

As we expect that the local codes far from the query point have small weights, we take the reciprocal of the cubic distance as the weight in \cref{eq:weight_}. Then we normalize the weights in \cref{eq:weight}.
After that, the matrix multiplication is applied to the weight matrix $\mathcal{W}$ and latent codes $\mathbf{C}$, which produces a matrix $\mathbf{Z}\in \mathbb{R}^{B \times m}$. Intuitively, each row of $\mathbf{Z}$ is a vector $\mathbf{z}_j \in \mathbb{R}^m$ for query point $\mathbf{q}_j$, which is interpolated from latent codes $\mathbf{C}$ based on distances. \cref{fig:interpolation} shows the interpolation process of our method, and the difference with traditional grid-based methods (e.g \cite{Peng2020ECCV}). At last, like most methods do, we leverage a multi-layer perceptron as the decoder. We concatenate $\mathbf{Q}$ and $\mathbf{Z}$ together as the input of decoder and get an output vector $\mathbf{d} \in \mathbb{R}^B$, in which each element $d_{j}$ is a predicted signed distance for $\mathbf{q}_j$.

As $\mathbf{p}_i$ can be any $(x,y,z)$ coordinates in the bounding box, our method is a more flexible representation whose code positions are continuous, while latent codes in other methods are usually constrained at regular and discrete positions. Moreover, we can dynamically update code positions during optimization directly by back-propagation and gradient descent, as there is no essential difference between $\mathbf{p}_i$ and other trainable parameters in the network.

\subsection{Code Position Loss}
\label{sec:code_position_loss}

To fully utilize the latent codes, we further propose a novel Code Position (CP) loss. Our motivation is to guide more latent codes to be distributed near the regions with complex geometric details. 

\begin{figure}[t]
    \centering
    \includegraphics[scale=0.8]{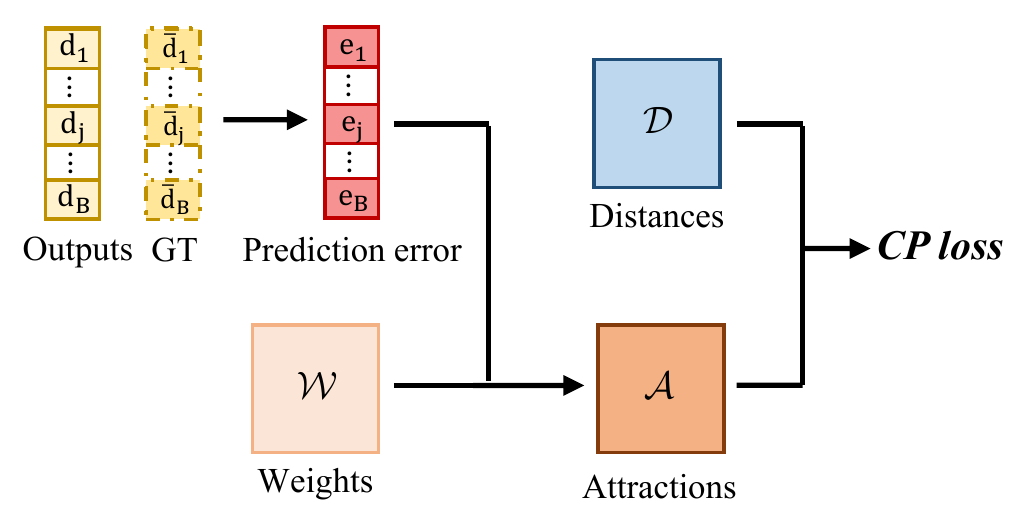}
    \caption{\textbf{Code Position Loss.} We assume query points are attractive to latent codes. And we define attractions based on prediction errors of query points. As elements of the weight matrix $\mathcal{W}$ in \cref{sec:dynamic_code_cloud} have the property of decreasing with increasing distances, we use $\mathcal{W}$ as a decay of attractions. Lastly, we apply the element-wise multiplication to the distance matrix and the attraction matrix, and take the average of all values as the final CP loss.}
    \label{fig:code_position_loss}
\end{figure}

As shown in \cref{fig:code_position_loss}, we first define the prediction error of each query point $\mathbf{q}_j$ as $e_j$, i.e.
\begin{equation}
    e_j = \mid d_j - \bar{d}_j \mid,
    \label{eq:prediction_error}
\end{equation}
where $d_j$ is the predicted signed distance for query point $\mathbf{q}_j$ by our method and $\bar{d}_j$ is the ground truth. Intuitively, larger $e_j$ means it is more difficult to reconstruct the local region near the corresponding query point, which further indicates that complex geometries details may exist on this region. As we expect the latent codes to get closer to the query points with higher prediction errors, we assume that there is a certain attraction force between query points and latent codes. Moreover, such attraction force should be directly proportional to $e_j$ and decay with the growth of distances between latent codes and query points. As elements of the weight matrix $\mathcal{W}$ in \cref{sec:dynamic_code_cloud} have the property of decreasing with increasing distances, we use $\mathcal{W}$ as a decay of attraction force. Therefore, we define the \emph{attraction matrix} $\mathcal{A} \in \mathbb{R}^{B \times n}$ between query points $\mathbf{Q}$ and latent codes $\mathbf{C}$ as
\begin{equation}
    \mathcal{A}_{ji} = e_j * \mathcal{W}_{ji}.
    \label{eq:attraction}
\end{equation}
Then, we apply element-wise multiplication between the attraction matrix $\mathcal{A}$ and the distance matrix $\mathcal{D}$, and take the average of all elements as the final Code Position Loss $L_{CP}$, denoted by
\begin{equation}
    L_{CP} = \frac{1}{B*n}\sum_{j=1}^B \sum_{i=1}^n \mathcal{A}_{ji} * \mathcal{D}_{ji}.
    \label{eq:code_position_loss}
\end{equation}
Note that we cut off gradient back propagation to $\mathcal{A}$. Thus, the distances between latent codes and query points are optimized based on the attraction, leading to further update of code positions $\mathbf{P}$.

With the guidance of CP loss, more latent codes will be distributed near the regions with complex geometric details, which enables our method to capture fine local geometric details. On the other hand, since there are fewer latent codes around simple geometric regions, this allows our method to represent a 3D shape with a small amount of latent codes, compared with previous grid-based methods \cite{Chibane2020ImplicitFI,Chen_2021_ICCV,Peng2020ECCV,Takikawa2021NeuralGL}. Although other methods can also assign more latent codes to complex regions, such as by refining the depth of octree \cite{Martel2021ACORNAC}, our method is more flexible and effective since our code positions are continuous. Furthermore, we optimize the code positions directly by back-propagation and gradient descent, which is more efficient.

\subsection{Training}
\label{sec:training}

We train our network in an auto-decoder \cite{Tan1995ReducingDD} framework. To optimize latent codes and code positions, sampled query points and their ground truth signed distances should be provided as training data. For fair comparisons, we adopt different sampling schemes in different experiments to keep the same setting with compared methods.

To optimize latent codes, we minimize the mean squared error (MSE) between predicted signed distances $d_j$ and ground truth $\bar{d}_j$, denoted as
\begin{equation}
    L_{MSE} = \frac{1}{B}\sum_{j=1}^B||d_j - \bar{d}_j||^2.
    \label{eq:l2_loss}
\end{equation}
We also minimize the CP loss to optimize code positions. As a result, our final loss $L$ is defined as
\begin{equation}
    L = L_{MSE} + \lambda L_{CP},
    \label{eq:loss}
\end{equation}
where $\lambda$ is a parameter to balance $L_{MSE}$ and $L_{CP}$.

\section{Experiments}
\label{sec:experiments}

In this section, we conduct experiments to evaluate the performance of DCC-DIF. Specifically, in \cref{sec:single_shape_fitting}, we demonstrate the ability of DCC-DIF for describing geometric details through the single shape fitting task. In \cref{sec:reconstructing_3d_datasets}, the ability of DCC-DIF for learning shape priors and generalizing to new objects is further evaluated, by applying DCC-DIF to reconstruct unseen shapes. In \cref{sec:ablation_study}, we validate the effects of each part of DCC-DIF. Limited by space, more discussions can be found in appendix.

\subsection{Single Shape Fitting}
\label{sec:single_shape_fitting}

We apply DCC-DIF to single shape fitting task to evaluate the ability of describing geometric details. In this experiment, the latest NGLOD \cite{Takikawa2021NeuralGL} is typically selected for our comparison, which is an octree based method and achieves state-of-the-art results in single shape fitting.

\begin{table*}[t]
  \centering
  \resizebox{14cm}{!}{
  \begin{tabular}{@{}l||cccccccc@{}}
    \toprule
    \multirow{2}*{Metrics} & \multicolumn{8}{c}{Methods} \\
    & DeepSDF \cite{Park2019DeepSDFLC} & FFN \cite{tancik2020fourfeat} & SIREN \cite{sitzmann2019siren} & NI \cite{Davies2020OverfitNN} & NGLOD3 \cite{Takikawa2021NeuralGL} & NGLOD4 \cite{Takikawa2021NeuralGL} & NGLOD5 \cite{Takikawa2021NeuralGL} & Ours \\
    \midrule
    IoU $\uparrow$ & 96.8 & 97.7 & 95.1 & 96.0 & 99.0 & 99.3 & 99.4 & \textbf{99.5} \\
    CD $\downarrow$ & $-$ & $-$ & $-$ & $-$ & 3.69 & 3.59 & 3.57 & \textbf{3.55} \\
    \textit{\#Codes} & $-$ & $-$ & $-$ & $-$ & 5.7K/0.9K & 41.7K/3.7K & 316K/15K & 5.6K \\
    \textit{\#Param.} & 1.8M & 527K & 264K & 7.6K & 4.7K & 4.7K & 4.7K &  4.7K \\
    \bottomrule
  \end{tabular}
  }
  \caption{Results on Thingi32 \cite{Zhou2016Thingi10KAD}. We compare the reconstruction quality and efficiency between our method and others (NGLOD \cite{Takikawa2021NeuralGL} with LODs equals to 3, 4 and 5 is denoted as NGLOD3, NGLOD4 and NGLOD5, respectively). For quality, we use the IoU and CD as metrics. For efficiency, we use \textit{\#Codes} and \textit{\#Param.} as metrics. The \textit{\#Codes} indicates the number of latent codes used in each method. And the \textit{\#Param.} means the number of network parameters used for a single distance query. In the \textit{\#Codes} row, for each LODs of NGLOD \cite{Takikawa2021NeuralGL}, we present the average number of latent codes before/after removing the empty nodes of octree. Our method simultaneously achieves the best quality and a high efficiency. }
  \label{tab:single_shape_fitting_results}
\end{table*}

\paragraph{Network settings.} For fair comparisons, we use the same settings with NGLOD \cite{Takikawa2021NeuralGL}. Specifically, we set the decoder to be a multi-layer perceptron with only one hidden layer, which is 128-dimensional and leverages a ReLU \cite{Glorot2011DeepSR} activation function. And we set $m$ to 32, which is the same with NGLOD \cite{Takikawa2021NeuralGL}. NGLOD \cite{Takikawa2021NeuralGL} has different numbers of latent codes for different shapes after removing the empty nodes of octree that contain no surface. We simply use $n = 5600$ latent codes for each shape, which is approximately equal to the number of latent codes used in NGLOD \cite{Takikawa2021NeuralGL} with 3 LODs (before removing the empty nodes of octree). As prediction errors $e_j$ and distances $\mathcal{D}_{ji}$ tend to be small, we choose $ \lambda $ as a relatively large number to balance $L_{MSE}$ and $L_{CP}$, where $\lambda$ is typically set to $7000$ in this experiment.

\paragraph{Data and metrics.} Following NGLOD \cite{Takikawa2021NeuralGL}, we also select the same 32 shapes from Thingi10K \cite{Zhou2016Thingi10KAD} and follow the same preprocessing practice as NGLOD. Specifically, following DualSDF \cite{Hao2020DualSDFSS}, we normalize the meshes and remove internal triangles. And we sign the distances with ray stabbing \cite{Nooruddin2003SimplificationAR}. We adopt the same schemes with NGLOD \cite{Takikawa2021NeuralGL} to obtain a point set for training. Specifically, we sample 500K points at each epoch, in which 100K points are sampled uniformly in the bounding box, 200K points from the object surface and the other points are sampled near the object surface. For metrics, we evaluate results using Chamfer Distance (CD) and Intersection over Union (IoU). Following NGLOD \cite{Takikawa2021NeuralGL}, we also pay attention to the efficiency of storage and computation. Here, the number of latent codes used in each method is denoted as \textit{\#Codes}, which roughly shows the storage cost. The number of network parameters used for a single distance query is denoted as \textit{\#Param.}, which roughly indicates the computation cost.

\begin{figure}
    \centering
    \includegraphics[scale=0.43]{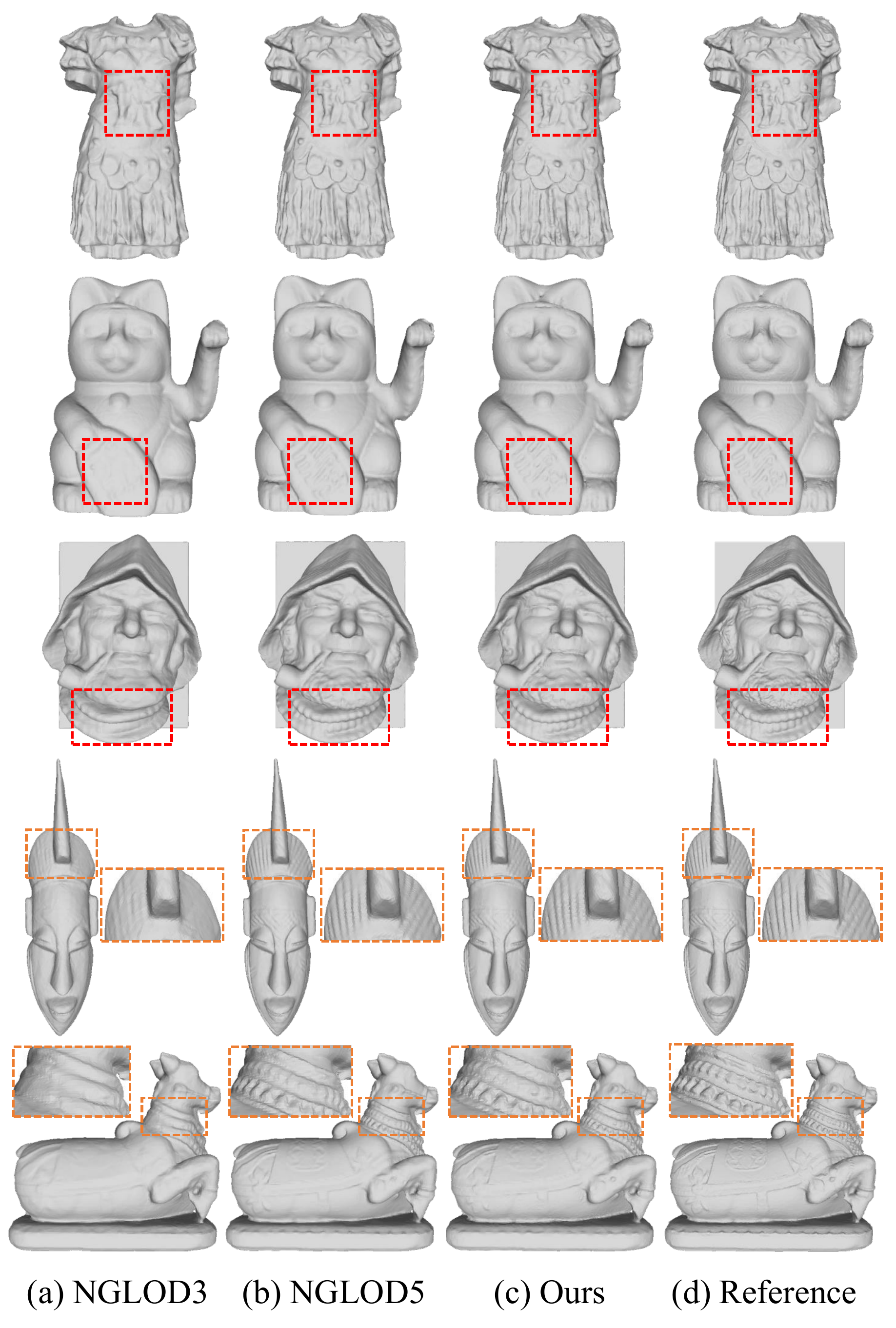}
    \caption{Visualization results on Thingi32 \cite{Zhou2016Thingi10KAD}. We visually compare with NGLOD \cite{Takikawa2021NeuralGL} with LODs that equals to 3 and 5, as denoted by NGLOD3 and NGLOD5, respectively. We achieve better results than NGLOD3, especially for local geometric details. Although NGLOD5 can achieve the similar results with our method, our method has a small number of latent codes than NGLOD5.}
    \label{fig:thingi32}
\end{figure}

\begin{table*}[t]
  \centering
  \resizebox{16cm}{!}{
  \begin{tabular}{@{}l||cccccc||cccccc@{}}
    \toprule
    \multirow{2}*{Category} & \multicolumn{6}{c}{Chamfer($\downarrow$)} & \multicolumn{6}{c}{F-Score($\uparrow$,\%)} \\
    & Occ. \cite{Mescheder2019OccupancyNL} & SIF \cite{Genova2019LearningST} & LDIF \cite{Genova2020LocalDI} & IF. \cite{Chibane2020ImplicitFI} & MDIF \cite{Chen_2021_ICCV} & Ours & Occ. \cite{Mescheder2019OccupancyNL} & SIF \cite{Genova2019LearningST} & LDIF \cite{Genova2020LocalDI} & IF. \cite{Chibane2020ImplicitFI} & MDIF \cite{Chen_2021_ICCV} & Ours \\
    \midrule
    airplane & 0.16 & 0.44 & 0.10 & 0.52 & 0.028 & \textbf{0.011} & 87.8 & 71.4 & 96.9 & 94.4 & 98.6 & \textbf{99.7} \\
    bench & 0.24 & 0.82 & 0.17 & 0.31 & 0.052 & \textbf{0.017} & 87.5 & 58.4 & 94.8 & 92.6 & 96.0 & \textbf{99.5} \\
    cabinet & 0.41 & 1.10 & 0.33 & 0.11 & \textbf{0.051} & 0.131 & 86.0 & 59.3 & 92.0 & 93.0 & \textbf{96.6} & 96.4 \\
    car & 0.61 & 1.08 & 0.28 & 0.30 & \textbf{0.088} & 0.218 & 77.5 & 56.6 & 87.2 & 87.4 & \textbf{93.0} & 92.7 \\
    chair & 0.44 & 1.54 & 0.34 & 0.10 & \textbf{0.035} & 0.037 & 77.2 & 42.4 & 90.9 & 94.5 & 97.6 & \textbf{99.1} \\
    display & 0.34 & 0.97 & 0.28 & 0.07 & \textbf{0.019} & 0.028 & 82.1 & 56.3 & 94.8 & 96.1 & 98.7 & \textbf{99.4} \\
    lamp & 1.67 & 3.42 & 1.80 & 1.17 & 0.795 & \textbf{0.327} & 62.7 & 35.0 & 84.0 & 89.1 & 93.5 & \textbf{97.3} \\
    rifle & 0.19 & 0.42 & 0.09 & 1.07 & 0.057 & \textbf{0.007} & 86.2 & 70.0 & 97.3 & 93.5 & 96.9 & \textbf{99.9} \\
    sofa & 0.30 & 0.80 & 0.35 & 0.13 & 0.037 & \textbf{0.036} & 85.9 & 55.2 & 92.8 & 92.5 & 98.4 & \textbf{99.1} \\
    speaker & 1.01 & 1.99 & 0.68 & 0.14 & \textbf{0.044} & 0.146 & 74.7 & 47.4 & 84.3 & 90.2 & \textbf{97.3} & 96.1 \\
    table & 0.44 & 1.57 & 0.56 & 0.17 & 0.046 & \textbf{0.029} & 84.9 & 55.7 & 92.4 & 93.4 & 97.6 & \textbf{99.3} \\
    telephone & 0.13 & 0.39 & 0.08 & 0.08 & \textbf{0.010} & 0.027 & 94.8 & 81.8 & 98.1 & 98.8 & \textbf{99.6} & 99.3 \\
    watercraft & 0.41 & 0.78 & 0.20 & 0.90 & 0.067 & \textbf{0.042} & 77.3 & 54.2 & 93.2 & 92.7 & 97.2 & \textbf{98.3} \\
    \midrule
    mean & 0.49 & 1.18 & 0.40 & 0.39 & 0.102 & \textbf{0.081} & 81.9 & 59.0 & 92.2 & 92.9 & 97.0 & \textbf{98.2} \\
    \bottomrule
  \end{tabular}
  }
  \caption{Results on ShapeNet \cite{Chang2015ShapeNetAI}. We use the Chamfer distance (CD) and F-Score to evaluate the reconstruction results of our and compared methods. Our method achieves the lowest mean CD and the highest mean F-Score, outperforming all other methods.}
  \label{tab:shapenet_results}
\end{table*}

\paragraph{Results and analyses.} 

\cref{tab:single_shape_fitting_results} shows the result comparison of our method and other methods, including DeepSDF \cite{Park2019DeepSDFLC}, FFN \cite{tancik2020fourfeat}, SIREN \cite{sitzmann2019siren}, NI \cite{Davies2020OverfitNN} and NGLOD \cite{Takikawa2021NeuralGL}. Specially, the LODs of NGLOD are selected as 3, 4 and 5, denoted as NGLOD3, NGLOD4 and NGLOD5, respectively. As we have the exactly same experiment settings with NGLOD \cite{Takikawa2021NeuralGL}, some results in the table directly come from it. Since the CD can be influenced by some factors such as the number of surface points, we recompute the CD by ourselves. The results in \cref{tab:single_shape_fitting_results} show that our method achieves the highest IoU and lowest CD, which are beyond other methods. We show the average number of latent codes used by NGLOD \cite{Takikawa2021NeuralGL} before/after removing the empty nodes of octree. It is worth noting that, compared with NGLOD5, our method leverages a small number of latent codes, but still achieves slightly better results. This demonstrates that our method can represent 3D shapes more efficiently. Our method also has advantages in storage and computation efficiency. As visualized in \cref{fig:thingi32}, our method achieves the similar reconstruction quality compared with NGLOD5, while using a small number of latent codes. Compared with NGLOD3, our method achieves better reconstruction quality, especially for local geometric details. 

\subsection{Reconstructing 3D Datasets}
\label{sec:reconstructing_3d_datasets}

We conduct an experiment to reconstruct 3D datasets using our method. Specifically, we optimize the latent codes, code positions and decoder parameters in the training phase. During inference, we fix decoder parameters and only optimize the latent codes and code positions on unseen shapes. This experiment shows the ability of our method to learn shape priors and generalize to new objects.

\paragraph{Network settings.} We leverage a decoder with the same structure as IM-Net \cite{Chen2019LearningIF}, which is a fully-connected network with connections between layers. We set $m=32$ and $n=1376$, thus we have the same number of parameters in latent codes with MDIF \cite{Chen_2021_ICCV}. To balance $L_{MSE}$ and $L_{CP}$, we set $\lambda = 3000$ in this experiment.

\paragraph{Data and metrics.} We use a subset of 13 categories in ShapeNet \cite{Chang2015ShapeNetAI} and divide the dataset with train/test splits from 3D-R$^2$N$^2$ \cite{Choy20163DR2N2AU}. And we generate watertight meshes with tools from OccNet \cite{Mescheder2019OccupancyNL}. In this experiment, we sample a point set from each shape, and use the same point set for all epochs during training. Each point set contains 200K samples where half of the points comes from uniform sampling and the others are sampled near the object surface. We use Chamfer L2 distance and F-Score as the metrics, which have the identical settings with LDIF and MDIF \cite{Genova2020LocalDI,Chen_2021_ICCV}.

\paragraph{Results and analyses.}

\begin{figure}
    \centering
    \includegraphics[scale=0.42]{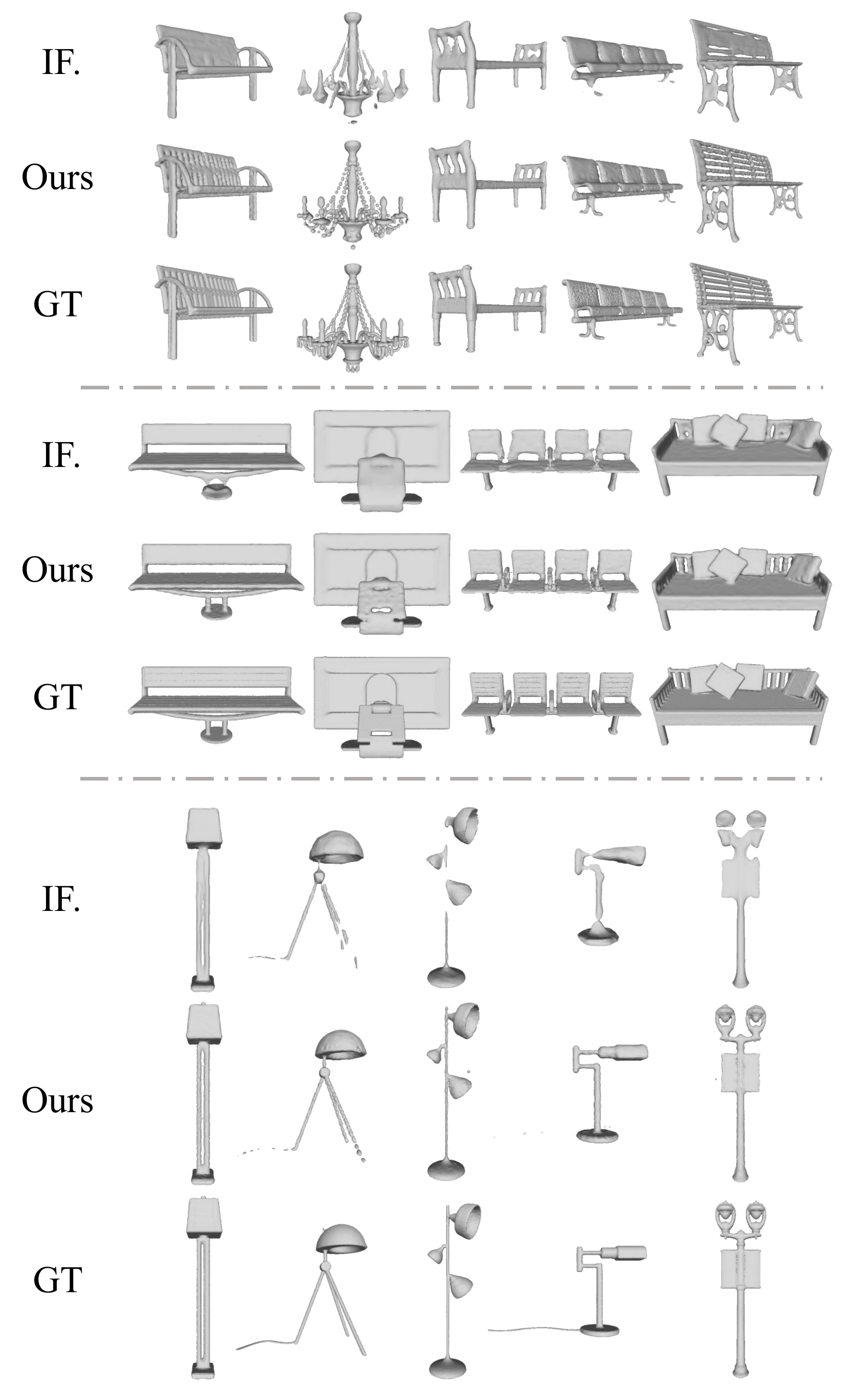}
    \caption{Visualization on ShapeNet \cite{Chang2015ShapeNetAI}. Compared to IF-Net \cite{Chibane2020ImplicitFI}, our method achieves better reconstruction quality in local regions with complex geometric details, such as thin strips and holes.}
    \label{fig:shapenet}
\end{figure}

The results are shown in \cref{tab:shapenet_results}. As we keep exactly the same experiment settings with MDIF \cite{Chen_2021_ICCV}, some results in the table directly come from it. Our method surpasses all other methods both on the mean CD and the mean F-Score, which demonstrates the ability of our method to learn shape priors and generalize to new objects. Among different categories of ShapeNet \cite{Chang2015ShapeNetAI}, there are great differences in difficulty of reconstruction. Shapes in some categories tend to be complex and various, such as lamp and rifle. In contrast, shapes in some other categories are relatively simple and similar to each other, such as speaker and display. From \cref{tab:shapenet_results}, we find that our method has great advantages in reconstructing complex and various shapes. \cref{fig:shapenet} visually shows the quality of our reconstruction on ShapeNet \cite{Chang2015ShapeNetAI}, compared with IF-Net \cite{Chibane2020ImplicitFI}. Our method achieves better reconstruction results, especially in local regions with thin strips and holes. This demonstrates that our method has a ability to represent complex shapes and capture fine local geometry details.

\subsection{Ablation Study}
\label{sec:ablation_study}

Among existing DIF methods, grid and octree based methods achieve good performance in both single shape fitting and 3D datasets reconstruction. Compared with these methods, there are three differences in DCC-DIF, including interpolation process, the novel position vectors and CP loss. To evaluate the effects of each difference, we conduct ablation studies on display and watercraft categories from ShapeNet \cite{Chang2015ShapeNetAI}, where display category contains relatively simple shapes and watercraft category tend to be complex. We keep other experiment settings the same as \cref{sec:reconstructing_3d_datasets}.

To evaluate the influence of interpolation process, we design two variations of DCC-DIF. The first variation fixes all latent codes to be located at grid points and applies trilinear interpolation. The second variation also fixes latent codes at grid points but leverages distance-based interpolation, as described in \cref{sec:dynamic_code_cloud}. As positions of latent codes are fixed, we remove the CP loss from both variations. As shown in \cref{tab:ablation_interpolation}, the variation using trilinear interpolation achieves better results. It demonstrates that the performance of our DCC-DIF does not benefit from the new interpolation algorithm, but from our proposed position vectors and CP loss.

\begin{table}
  \centering
  \resizebox{5cm}{!}{
  \begin{tabular}{@{}l||cc||cc@{}}
    \toprule
    \multirow{2}*{Category} & \multicolumn{2}{c}{Chamfer($\downarrow$)} & \multicolumn{2}{c}{F-Score($\uparrow$,\%)} \\
    & Trilinear & Ours & Trilinear & Ours \\
    \midrule
    display & 0.038 & 0.045 & 99.2 & 98.8 \\
    watercraft & 0.090 & 0.108 & 97.2 & 96.3 \\
    \bottomrule
  \end{tabular}
  }
  \caption{Comparison of interpolation process.}
  \label{tab:ablation_interpolation}
\end{table}

Next, we validate the effects of our proposed position vectors and CP loss. As a baseline, we remove the CP loss from our full version pipeline and fix all latent codes at grid points. Then we evaluate the benefit of position vectors and CP loss respectively by two variations of our DCC-DIF. The first variation only adds the position vectors to the baseline and the second variation adds both position vectors and CP loss to baseline. Results are shown in \cref{tab:ablation_pv_cp}. We can find that both position vectors and the CP loss play a positive role in 3D shape representation, which supports our proposal. Additionally, the CP loss shows more significant effects with complex shapes, which is consistent with our design to guide more latent codes to be distributed around complex geometric details.

\begin{table}
  \centering
  \resizebox{8cm}{!}{
  \begin{tabular}{@{}l||ccc||ccc@{}}
    \toprule
    \multirow{2}*{Category} & \multicolumn{3}{c}{Chamfer($\downarrow$)} & \multicolumn{3}{c}{F-Score($\uparrow$,\%)} \\
    & baseline & p.v. & p.v. + CP & baseline & p.v. & p.v. + CP \\
    \midrule
    display & 0.045 & 0.033 & 0.028 & 98.8 & 99.3 & 99.4 \\
    watercraft & 0.108 & 0.081 & 0.042 & 96.3 & 97.8 & 98.3 \\
    \bottomrule
  \end{tabular}
  }
  \caption{Ablation of position vectors and CP loss. The 'p.v.' denotes position vectors.}
  \label{tab:ablation_pv_cp}
\end{table}

\section{Conclusion and Limitation}
\label{sec:conclusion}

In this paper, we introduce a novel DCC-DIF to learn deep implicit functions for 3D shapes. Among existing DIF methods, the best results are achieved by grids or octree based methods. However, the latent codes in these methods are constrained to be located at discrete and regular positions, and the code positions are difficult to be optimized. In contrast, the code positions in our DCC-DIF are continuous and flexible by explicitly assigning a position vector to each latent code. We further propose a novel CP loss to optimize the positions of latent codes, so that more latent codes are distributed around complex geometric details. In experiments, our method outperforms other methods and achieves state-of-the-art results, which demonstrates its performance and efficiency. The ablation studies show effects of each part of our design, which supports our proposal.

Some limitations exist in our DCC-DIF, which are also the directions of improvement in our future work. First, the current DCC-DIF is unable to represent different levels of details like NGLOD\cite{Takikawa2021NeuralGL}. To address this problem, we plan to design a hierarchical network, in which each level leverages a DCC-DIF and the number of latent codes growth along with the increase of level. Another limitation is that current DCC-DIF is unsuited to global operations such as completion \cite{Dai2017ShapeCU}. 
Inspired by MDIF \cite{Chen_2021_ICCV}, we can further set the first level of above-mentioned hierarchical DCC-DIF to be a single global latent code and design a novel module for information exchange between global and local codes.

{\small
\bibliographystyle{ieee_fullname}
\bibliography{references}
}

\end{document}